\documentclass[sigconf]{acmart}

\usepackage{latexsym}
\usepackage{algorithmic}
\usepackage{amsmath}
\usepackage{esvect}
\usepackage{latexsym}
\usepackage{boxedminipage} 
\usepackage{CJK}
\usepackage{enumitem}
\usepackage{mathrsfs}
\usepackage{amsfonts}
\usepackage{bm}
\usepackage{colortbl}
\usepackage{graphicx}
\usepackage{float}
\usepackage{subfigure}
\usepackage{soul}
\usepackage{booktabs} 
\usepackage{url}
\usepackage{amsmath}
\usepackage{xcolor}
\usepackage{algorithm}
\usepackage{makecell}
\usepackage{xspace}
\usepackage{array}
\usepackage[normalem]{ulem}
\usepackage{footmisc}
\usepackage{colortbl}
\usepackage{epstopdf}
\usepackage{setspace}
\usepackage{pifont}
\usepackage{rotating,tabularx}
\usepackage{multirow}

\definecolor{gred}{RGB}{219,68,55}
\definecolor{gblue}{RGB}{66,133,244}
\definecolor{gyellow}{RGB}{244,180,0}
\definecolor{ggreen}{RGB}{15,157,88}
\definecolor{ggrey}{RGB}{115,115,115}

\newcommand{\ie}{\emph{i.e.,}\xspace}
\newcommand{\error}[1]{\textcolor{gred}{\textbf{#1}}} 
\newcommand{\fph}[1]{\textcolor{gblue}{\textbf{#1}}} 
\newcommand{\reph}[1]{\textcolor{ggreen}{\textbf{#1}}} 
\newcommand{\specialcell}[2][c]{%
	\begin{tabular}[#1]{@{}c@{}}#2\end{tabular}}
\AtBeginDocument{%
  \providecommand\BibTeX{{%
    \normalfont B\kern-0.5em{\scshape i\kern-0.25em b}\kern-0.8em\TeX}}}

\copyrightyear{2022}
\acmYear{2022}
\setcopyright{rightsretained}
\acmConference[SIGIR '22]{Proceedings of the 45th International ACM SIGIR Conference on Research and Development in Information Retrieval}{July 11--15, 2022}{Madrid, Spain.}
\acmBooktitle{Proceedings of the 45th Int'l ACM SIGIR Conference on Research and Development in Information Retrieval (SIGIR '22), July 11--15, 2022, Madrid, Spain}
\acmDOI{10.1145/3477495.3532065}
\acmISBN{978-1-4503-8732-3/22/07}

\usepackage{etoolbox}

\makeatletter
\patchcmd{\maketitle}{\@copyrightpermission}{
   \begin{minipage}{0.3\columnwidth}
     \href{http://creativecommons.org/licenses/by/4.0/}{\includegraphics[width=0.70\textwidth]{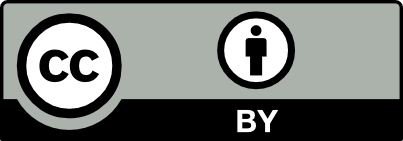}}
   \end{minipage}\hfill
   \begin{minipage}{0.7\columnwidth}
     \href{http://creativecommons.org/licenses/by/4.0/}{This work is licensed under a Creative Commons Attribution International 4.0 License.}
   \end{minipage}
  
}{}{}

\makeatother

\begin{document}

\fancyhead{}
\title{Target-aware Abstractive Related Work Generation with Contrastive Learning}

\author{Xiuying Chen}
\affiliation{%
  \institution{CBRC, KAUST\\
  CEMSE, KAUST}
}
\email{xiuying.chen@kaust.edu.sa}

\author{Hind Alamro}
\affiliation{%
  \institution{CBRC, KAUST\\
  CEMSE, KAUST}
}
\email{hind.alamro@kaust.edu.sa}

\author{Mingzhe Li}
\affiliation{%
  \institution{Ant Group}
}
\email{li_mingzhe@pku.edu.cn}

\author{Shen Gao}
\affiliation{%
  \institution{Peking University}
}
\email{shengao@pku.edu.cn}

\author{Rui Yan$^{\dagger}$}
\affiliation{%
  \institution{Renmin University of China}
}
\email{ruiyan@ruc.edu.cn}

\author{ Xin Gao}
\affiliation{%
  \institution{CBRC, KAUST\\
  CEMSE, KAUST}
}
\email{xin.gao@kaust.edu.sa}

\author{Xiangliang Zhang$^{\dagger}$}
\affiliation{%
  \institution{University of Notre Dame\\CEMSE, KAUST
  }
}
\email{xzhang33@nd.edu}

\def\authors{Xiuying Chen, Hind Alamro, Mingzhe Li, Shen Gao, Rui Yan, Xin Gao, Xiangliang Zhang}

\thanks{$\dagger$ Corresponding authors.} 


\begin{abstract}
 
The related work section is an important component of a scientific paper, which highlights the contribution of the target paper in the context of the reference papers.
Authors can save their time and effort by using the automatically generated related work section as a draft to complete the final related work.
Most of the existing related work section generation methods rely on \textit{extracting} off-the-shelf sentences to make a comparative discussion about the target work and the reference papers.
However, such sentences need to be written in advance and are hard to obtain in practice.
Hence, in this paper, we propose an \textit{abstractive} target-aware related work generator (TAG), which can generate related work sections consisting of new sentences.
Concretely, we first propose a target-aware graph encoder, which models the relationships between reference papers and the target paper with target-centered attention mechanisms.
In the decoding process, we propose a hierarchical decoder that attends to the nodes of different levels in the graph with keyphrases as semantic indicators.
Finally, to generate a more informative related work, we propose multi-level contrastive optimization objectives, which aim to maximize the mutual information between the generated related work with the references and minimize that with non-references.
Extensive experiments on two public scholar datasets show that the proposed model brings substantial improvements over several strong baselines in terms of automatic and tailored human evaluations\footnote{\url{https://github.com/iriscxy/Target-aware-RWG}}.
\end{abstract}



\maketitle

\section{Introduction}

\begin{figure}[h]
	\centering
	\includegraphics[width=1\linewidth]{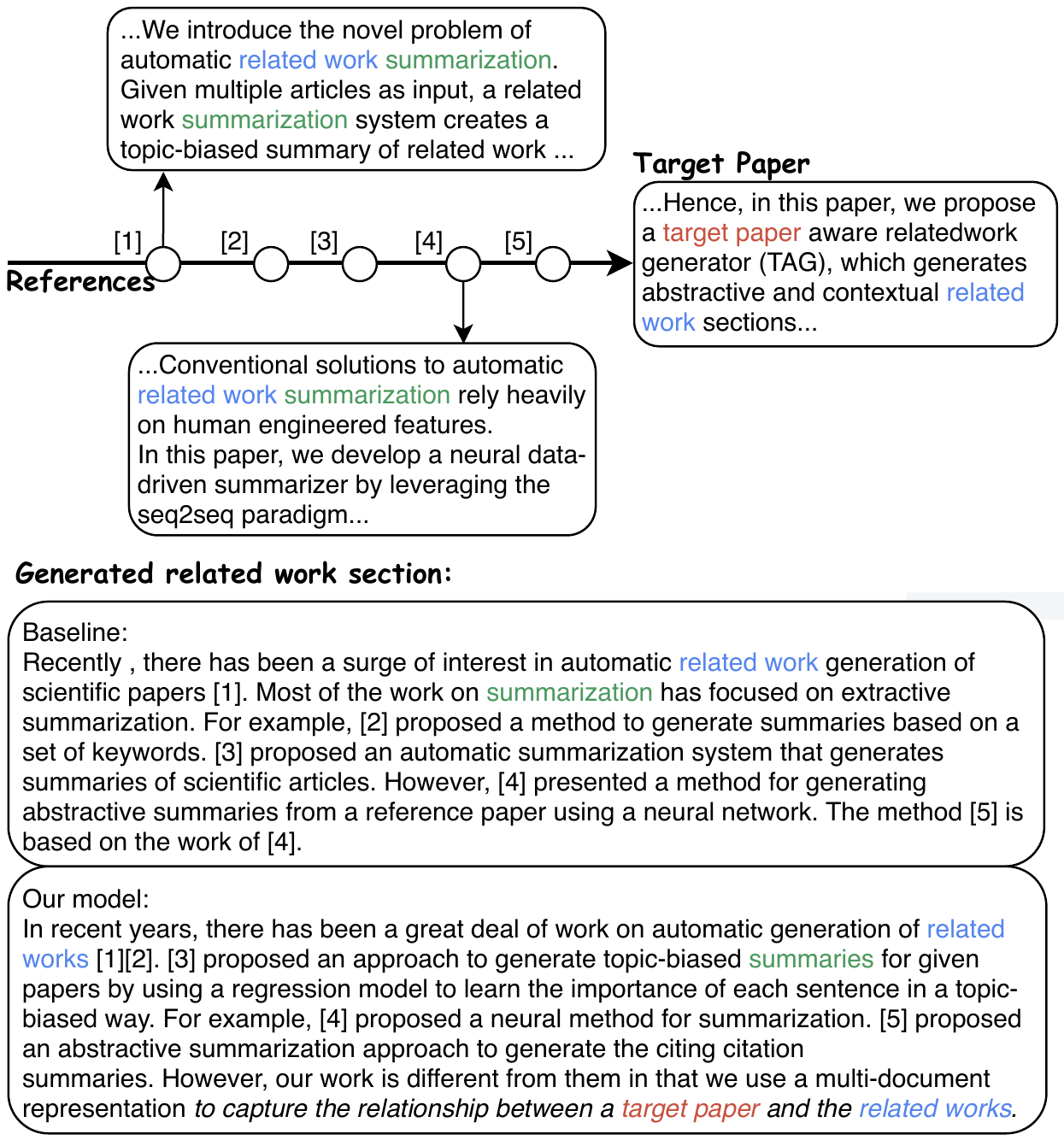}
	\caption{
	Related work section   generated by the baseline (RRG) and our model for this submission.
    Our model TAG successfully includes the keyphrase information and comparison of the references and the target paper (this submission), while the baseline model missed the comparison.
	}
	\label{fig_firstpage}
\end{figure}
Good scientific papers often need to help readers comprehend the contribution by contextualizing the proposed work in the scientific fields.
The related work section serves as a pivot for this goal, which compares and connects the innovation and novelty of the current work with previous studies.
The task of \textit{Related Work section Generation} (RWG) is thus proposed \cite{hoang2010towards}, which aims to generate a related work section for a target paper given multiple cited papers as reference.
The generated related work section can be used as a draft for the author to complete his or her final related work section.
Hence, RWG can greatly reduce the author's time and effort when writing a paper \cite{hu2014automatic}.

The unique characteristics of the task and the use of rare, technical terms in scientific articles make RWG challenging.
Conventional extractive RWG methods rely on off-the-shelf sentences to generate a target-aware related work section. 
For example, \citet{hu2014automatic} split the sentences of the references and the target paper into different topic-biased parts, and then apply importance scores to each sentence.
Subsequently, \citet{chen2016summarization} collected sentences from the target paper and papers that cite the reference papers to form a related work.
In such cases, sentences that make comparative discussion need to be written in advance, which still require human efforts and are hard to obtain.
Abstractive text generation technique is a natural solution to this problem, which has sophisticated abilities such as paraphrasing and generalization.
\citet{xing2020automatic} and \citet{ge2021baco} proposed to generate a short text to explain the connection between a pair of papers. However, they only considered sentences with a single citation, and their proposed model also hard-coded the setting of two input sources, instead of a variable number of input sources.
\citet{chen2021capturing} aimed to generate related work for a set of reference papers, but they didn't consider the target paper information.
Thus, the generated related work cannot describe how the target relates to the cited works.

To tackle the above problems and generate a more comprehensive related work, we propose a target-aware related work generator (TAG), which generates abstractive and contextual related work sections.
Concretely, we first propose a target-aware graph encoder, where the target paper interacts with the references papers under target-centered attention mechanisms.
Herein, the interaction consists of direct information flow between papers and indirect flow with keyphrases as an intermediary.
The former allows global information exchange, while the latter lets information interact under a specific topic. 
In the decoding phase, we introduce a hierarchical decoder that first attends to the keyphrase nodes in the graph and then projects the attention weights on the reference papers, and keyphrases.
In this way, we let the keyphrases be the semantic indicator and guide the generation process as a more abstract summarization clue.
To generate more informative related work sections with substantial content, we adapt the contrastive learning mechanism to the RWG task, where we aim to maximize the mutual information between the generated text with reference papers and minimize that with non-references.

We evaluate our model on two large-scale benchmark scholar datasets collected from S2ORC and Delve.
S2ORC consists of papers in multiple domains (physics, math, computer science, \textit{etc.}), and Delve consists of computer science papers.
We show that TAG outperforms the state-of-the-art summarization and RWG baselines on both datasets in terms of both ROUGE metrics and tailored human evaluations. 
Since our contrastive learning module can be an add-on module, we also demonstrate its effectiveness on various baseline models.
A part of the generated related work for this submission of our model and the most recent baseline is shown in Figure \ref{fig_firstpage}, illustrating the advantage of TAG over the baseline. 

Our contributions can be summarized as follows: 

$\bullet$ We first propose to incorporate the target paper information into the abstractive related work generation task with multiple citations.

$\bullet$ To solve this task, we propose a target-aware related work generator that first models the relationship between target paper and the cited papers by keyphrases, and then generates the related work guided by contrastive learning.
The generated related work can contextualize the target work in related fields.

$\bullet$ Experiments conducted on two benchmark datasets show that our model outperforms all the state-of-the-art approaches.
Human evaluations also demonstrate that most of the generated related works are more constructive than those produced by the baseline methods.

\section{Related Work}

The task of RWG is pioneered by \citet{hoang2010towards}, who proposed a prototype extractive summarization system taking keywords as input to drive the creation of an extractive summary.
Subsequently, \citet{hu2014automatic} proposed to split the sentences of the reference papers into different topic-biased parts for selection, and \citet{chen2016summarization} resorted to the papers that cite the reference papers of the writing paper and extracts the corresponding citation sentences to form a related work section.
\citet{wang2018neural,deng2021automatic} developed a data-driven neural summarizer by leveraging the seq2seq paradigm to extract sentences from references.
These methods are either unable to make a comparative discussion or rely on comparative sentences to be written in advance.
Such sentences still needed to be written by scientists and are hard to obtain in practice.

Due to the development of neural network-based models \cite{Gao2019How,Gao2020From,chen2021reasoning} for text generation, recent RWG focuses on generating new words and phrases to form related work.
For example, \citet{xing2020automatic} extended the Pointer-Generator Network \cite{See2017GetTT} to propose a model with a cross-attention mechanism that takes both the citation context and the cited paper's abstract as the input. 
\citet{ge2021baco} extended \cite{xing2020automatic} by encoding citation network information using a graph attention network in addition to a hierarchical bidirectional LSTM \cite{hochreiter1997long}.
However, these two works share the same limitation and are only applicable to citation texts with a single citation.
\citet{chen2021capturing} proposed an RWG model taking multiple citations as input, but did not consider the target paper information.
Thus, it still cannot generate a comparative related work discussion on the target paper.

The RWG task is similar to a multi-document summarization (MDS) task with multiple documents as input.
Classic MDS methods are extractive over graph-based representations of sentences.
For example, \citet{zhao2020summpip} converted the original documents to a sentence graph, taking both linguistic and deep representation into account, and \citet{liu2021unsupervised} constructed sentence graph based on both the similarities and relative distances in the neighborhood of each sentence.
Due to the success of neural abstractive models on single-document summarization, abstractive models are adapted to multi-document summarization. 
\citet{Jin2020MultiGranularityIN} proposed a multi-granularity interaction network for extractive and abstractive multi-document summarization, and \citet{zhou2021entity} augmented the classical Transformer \cite{vaswani2017attention} with a heterogeneous graph consisting of entity nodes.
However, MDS aims at synthesizing the similar and removing the redundant information among multiple input documents, while RWG needs to find the specific contributions of individual papers even if their research directions are the same and arrange them in comparative discussion \cite{hu2014automatic}.

\begin{figure*}
	\centering
	\includegraphics[width=0.9\linewidth]{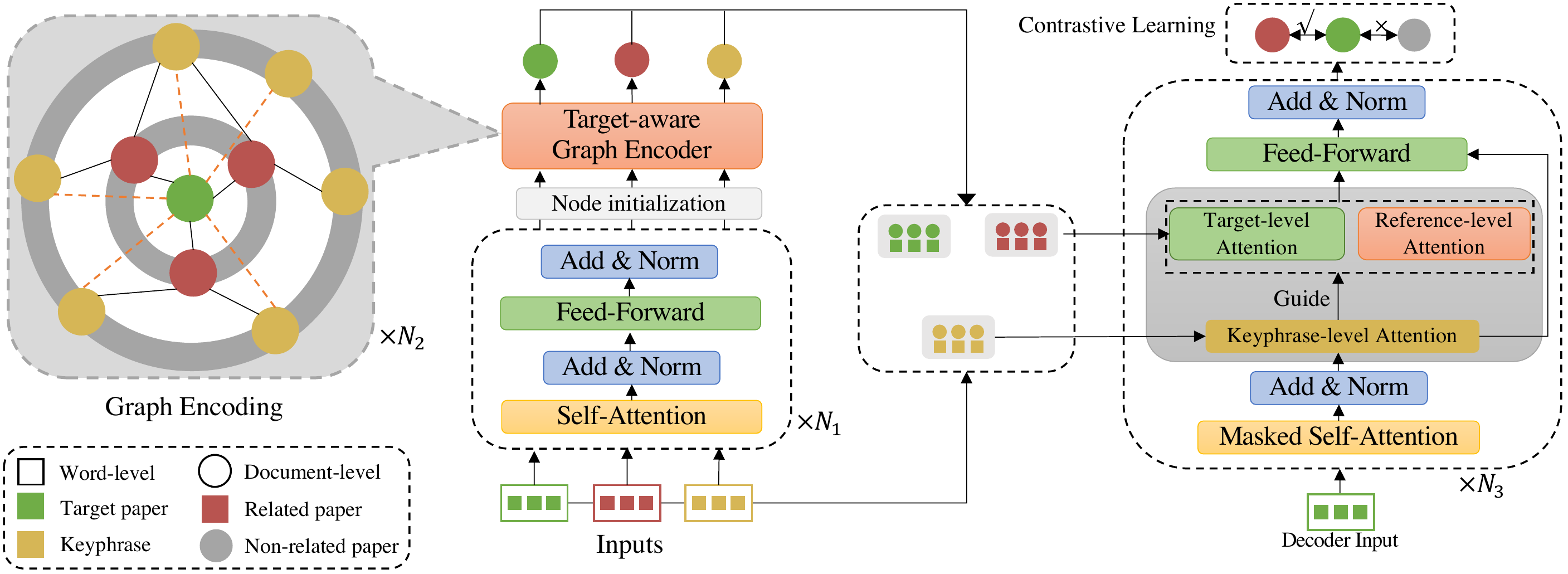}
	\caption{Overall architecture of our model, which consists of two parts:
	(1) \textit{Target-aware Graph Encoder} (middle and left) relates one paper to another and obtains contextual representations for nodes in different levels (circles in different colors).
    The node initialization and graph encoding are shown in the middle and the left part, respectively.
	(2) \textit{Hierarchical Decoder} (right) generates the related work by attending to multi-granular information in the graph guided by the keyphrases.
	}
	\label{fig:overview}
\end{figure*}

Our model is also related to contrastive learning, which has boosted unsupervised representation learning in recommendation \cite{qin2021world, zhang2021causerec}, computer vision \cite{li2020unimo,nan2021interventional}, and natural language processing \cite{zhu2021contrastive,mine1}. 
In text generation,
\citet{logeswaran2018efficient} proposed to learn better sentence representations by using a classifier to distinguish context sentences from other contrastive sentences.
\citet{bui2021self} introduced contrastive learning into code summarization, where the model is asked to recognize similar and dissimilar code snippets through contrastive learning objectives.
In scientific paper related studies,
\citet{cohan2020specter} explored contrastive learning and designed a loss function that trains the Transformer model to learn closer representations for papers when one cites the other.
The contrastive learning mechanism in our paper is also based on citation graph relatedness but is employed differently for generating comparative related work discussion.

\section{The Proposed TAG Model}

In this section, we first define the task of target-aware related work generation, then describe our TAG model in detail.

\subsection{Problem Formulation}

Given the abstract section of the target paper $A$ and its $|\mathcal{R}|$-size reference abstract collection $\mathcal{R} = \{r_1, ..., r_{|\mathcal{R}|}\}$,  we assume there is a ground truth related work $Y = (y_1, ..., y_{L_y})$.
To be specific, the target paper $A$ is represented as a sequence of $L_a$ words $(w_{0,1}, ..., w_{0,L_a})$, and each reference $r_i \in \mathcal{R}$ is $(w_{i,1}, ..., w_{i,L_r})$, where $i=1,...,|\mathcal{R}|$.
$L_a$ is the word number in the target paper, and $L_r$ is the word number in a reference.
We choose the abstract section to represent each input paper because the abstract section is a natural summarization of the corresponding paper while other sections may contain extreme details of the specific work \cite{hu2014automatic}.
Given $A$ and $\mathcal{R}$, our model generates a related work $\hat{Y} = \{\hat{y}_1, \hat{y}_2, ..., \hat{y}_{L_{\hat{y}}}\}$.
Finally, we use the difference between the generated $\hat{Y}$ and the ground truth $Y$ as the training signal to optimize the model parameters.

\subsection{Model Overview}
Our model is illustrated in Figure~\ref{fig:overview}, which follows the Transformer-based encoder-decoder architecture. 
We augment the encoder with a target-aware graph neural network, which models the relationship between
target and reference papers 1) \emph{directly} via their citation relations (red and green nodes in the middle of Figure~\ref{fig:overview}); and 2) \emph{indirectly} via the keyphrases of both target and reference papers (yellow nodes).
Correspondingly, we design a hierarchical decoder to attend to the multi-granularity information in the graph encoder guided by the keyphrases.
To generate an informative comparison to related papers, we also design a contrastive learning module to maximize the mutual information between the generated text with related papers and minimize that with unrelated papers.

\subsection{Target-aware Graph Encoder}
\label{sec:paper-encoder}

\noindent \textbf{Graph Construction.}
The related work section discusses the relations between the target paper and the related reference papers while encoding each paper separately cannot model the hierarchical structures and the relations that might exist among papers.
Hence, we design a target-aware graph encoder, where we model the relation between papers through different channels, \ie by direct edges and indirect edges through keyphrases.
By direct edges, we let information flow globally in the graph, while by keyphrase edges, we only establish edges for references that are related to the target paper under a specific key topic.
We choose the keyphrase as the basic semantic unit as it contains compact and salient information across different papers, whose representations can also be used to guide the decoding process later.

Concretely, the graph is constructed with the target paper node in the center. 
Reference papers are linked with the target paper, and keyphrases are linked with reference papers if the reference contains a keyword in the keyphrases.
All keyphrases are linked with the target paper via hyperlinks.
The intuition is that the generated related work serves for the target paper, thus, the target paper should play a central role in the graph.
Nodes in the same level are also pairwisely connected to enable information flow in each granularity, denoted by the circles in Figure~\ref{fig:overview}.

\noindent \textbf{Node Initialization.}
Transformer \cite{vaswani2017attention} is an encoder-decoder framework that could well capture the deep interaction between words in a document. 
Hence, we use Transformer to initialize the node representations in the graph.
 
We first employ a stack of $N_1$ token-level Transformer layers to obtain contextual word representations in each paper.
Each layer has two sub-layers: a self-attention mechanism and a fully connected feed-forward network.
We take the $i$-th reference paper to illustrate this process.

For the $l$-th Transformer layer, we first use a fully-connected layer to project document state $h^{l-1}_{w_{i,j}}$ into the query, \ie $Q^{l-1}_{i,j} = F_q(h^{l-1}_{w_{i,j}})$.
For self-attention mechanism, the key and value are obtained in a similar way: \ie $K^{l-1}_{i,j} = F_k(h^{l-1}_{w_{i,j}})$, $V^{l-1}_{i,j} = F_v(h^{l-1}_{w_{i,j}})$.
	Then, the attention sub-layer is defined as:
		\begin{align}
	\alpha_{i, j} =\frac{\exp \left(Q^{l-1}_{i,j} K^{l-1}_{i,j}\right)}{\sum_{n=1}^{L_r} \exp \left(Q^{l-1}_{i,n} K^{l-1}_{i,n}\right)}, 
	a_{i} =\sum_{j=1}^{L_r} \frac{\alpha_{i, j} V^{l-1}_{i,j}}{\sqrt {d_e}}, 
	\end{align}
	where $d_e$ stands for hidden dimension.
	The above process is summaized as $\text{MHAtt}(h^{l-1}_{w_{i,j}},h^{l-1}_{w_{i,*}})$, where $*$ denotes index from $1$ to $L_r$.
	Intuitively, the updated representation of $Q$ is formed by linearly combining the entries of $V$ with the weights.
	
	Then, a residual feed-forward sub-layer takes the output of self-attention sub-layer as the input:
\begin{align*}
\hat{h}_{w_{i,j}}^{l-1}&=\text{LN}\left(h_{w_{i, j}}^{l-1}+\text{MHAtt}\left(h_{w_{i, j}}^{l-1}, h_{w_{i,*}}^{l-1}\right)\right), \\
h_{w_{i,j}}^{l}&=\text {LN }\left(\hat{h}_{w_{i,j}}^{l-1}+\operatorname{FFN}\left(\hat{h}_{w_{i,j}}^{l-1}\right)\right),
\end{align*}
where FFN is a feed-forward network with an activation function, LN is a layer normalization \cite{ba2016layer}.

From the word-level representation we obtain the document-level representation $h_{d_i}$ for each reference paper by   mean-pooling.
Similarly, we obtain the word-level and paper-level representations for the target paper denoted as $h_{w_{0,*}}$ and $h_{d_{0}}$, respectively.

For the keyphrases, we utilize the keyword extraction tool KeyBERT \cite{grootendorst2020keybert} to extract keyphrases from both the target paper and reference papers, and use the same initialization method to obtain the keyphrase representations.
The word-level representation for the $i$-th keyphrase is denoted as $h_{w^c_{i,1},...,h_{w^c_{i,L_c}}}$, and the overall representation for $i$-th keyphrase is obtained by $h_{c_i}=\text{MeanPooling}(h_{w^c_{i,1},...,h_{w^c_{i,L_c}}})$, where $L_c$ is the word number in each keyphrase.
The total keyphrase number is denoted as $|\mathcal{C}|$.

\noindent \textbf{Graph Encoding.}
The graph encoding is a stack of $N_2$ identical layers, starting from the learning of keyphrase representations, then the reference paper representation, and last the target paper representation.
Concretely, the \textit{keyphrase} representations are obtained by performing multi-head self-attention and cross-attention on the constructed graph. 
The query in cross-attention is a keyphrase node representation, and the key and value are the adjacent paper node representations. 
The attention results are concatenated and processed by a feed-forward network, outputting $\hat{h}_{c_i}^l$ for one keyphrase node, where $l$ denotes the layer index.

\begin{figure}
    \centering
    \includegraphics[width=1\columnwidth]{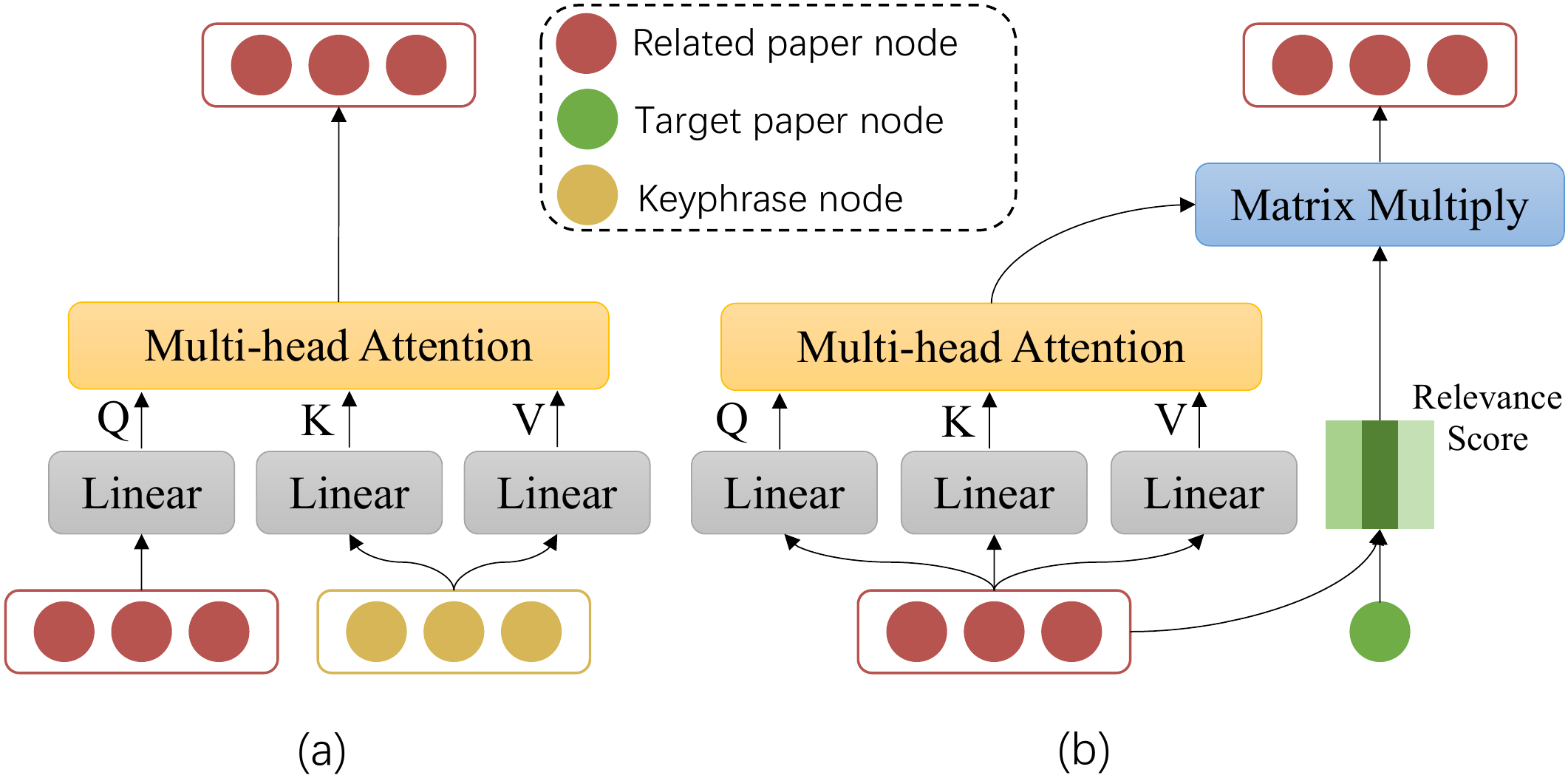}
    \caption{
          (a) Cross attention mechanism  for   reference papers and the keyphrases.
          (b) Target-centered attention mechanism, which captures semantic information within references under the guidance of the target paper.
    }
    \label{fig:attn}
\end{figure}

The \textit{reference} paper representation is updated from
three sources:
(1) performing multi-head self-attention across all references;
(2) performing multi-head cross-attention to obtain the keyphrase-aware reference paper representation. As shown in Figure~\ref{fig:attn}(a),  the reference representation is taken as the query, and the keyphrases representations are taken as the keys and values.
(3) performing our designed target-centered attention mechanism to obtain reference paper representations, as shown in Figure~\ref{fig:attn}(b). 
The target-centered attention starts with the application of self-attention mechanism on the reference papers:
\begin{align}
    \dot{h}_{d_{i}}^{l-1}&=\text{MHAtt}\left(h_{d_{i}}^{l-1}, h_{d_*}^{l-1}\right).
\end{align}
Then, taking the target paper information $h_{d_0}^{l-1}$ as condition, the attention score $ \beta_{i}^{l-1}$ on each original reference representation $h_{d_i}^{l-1}$ is calculated as:
\begin{align}
    \beta^{l-1}_{i} &=\text{softmax}\left(\text{FFN}\left(h_{d_i}^{l-1} (h_{d_0}^{l-1})^T\right)\right).
\end{align}
The target-aware reference representation is denoted as $\beta_{i}^{l-1} \dot{h}_{d_i}^{l-1}$.
In this way, we highlight the salient part of the references under the guidance of the target paper.
Last, a feed-forward network is employed to integrate three information sources to obtain the updated reference representation $\hat{h}_{d_{i}}^{l}$.

The \textit{target} paper representation is  {updated similarly with two cross-attention operations on references and keyphrases.
The attention results are concatenated and processed by a feed-forward layer, denoted as $\hat{h}^l_{d_0}$.}

Finally, we update word-level representations for references with the graph encoder outputs by a feed-forward network:
\begin{align}
    h_{w_{i,j}}=\text{FFN}(h^{N_1}_{w_{i,j}}+\hat{h}^{N_2}_{d_{i}}).
\end{align}
Word-level representation for the target paper $h_{w_{0,j}}$ and for the keyphrases $h_{c_{i,j}}$ are obtained in a similar way with the corresponding graph outputs.

\subsection{Hierarchical Decoder}

In the previous graph encoder, we obtain contextual and polished representations for the target paper, the references, and the keyphrases.
In the decoder part, we want to incorporate these information sources, where the target paper and the references are needed for generating a comparative related work.
Note that we also have the representations for the keyphrases, which are natural guidance for the generation process, as the keyphrases can be viewed as a special type of summarization that ``summarizes'' texts into a more abstract format.
Hence, we design a hierarchical decoder that firstly focuses on the keyphrases and then attends to the target paper and the other references. 
Our hierarchical decoder also follows the style of Transformer, consisting of $N_3$ identical layers.

For each layer, at the $t$-th decoding step, we apply the self-attention on the masked summary embeddings ($\text{MMHAtt}$), obtaining $g_t$.
The masking mechanism ensures that the prediction of the position $t$ depends only on the output of the position before $t$.
\begin{align*}
    g^{l}_t=\text {LN }\left(g^{l-1}_t+\text{MMHAtt}\left(g^{l-1}_t, g^{l-1}_*\right)\right).
\end{align*}
Based on $g_t^l$ we compute the context vectors over keyphrases using cross-attention multi-head attention (MHAtt):
\begin{align}
c^l_{c,t}=\text{MHAtt}\left(g^l_t, h_{c_*}\right).
\end{align}
We then use the keyphrase context vectors to guide the attention on other two-level nodes.
Take the target paper for example,
another MHAtt layer is applied to select relevant parts of word sequence from the target paper:
\begin{align}
c^l_{a,t}=\text{MHAtt}\left(c^l_{c,t}, h_{w_{0,*}}\right).
\end{align}
In a similar way, we obtain the context vectors $c^l_{r,t}$ on references.

These context vectors, treated as salient contents summarized from various sources, are concatenated with the decoder hidden state $g_t^{N_3}$ to produce the distribution over the target vocabulary:
\begin{align}
    P_t^{\text {vocab }}=\text{Softmax}\left(W_{o}\left[g^{N_3}_{t} ; c^{N_3}_{a,t};c^{N_3}_{r,t};c^{N_3}_{c,t}\right]\right),
\end{align}
where $[;]$ denotes concatenation operation.
All the learnable parameters are updated by optimizing the negative log likelihood objective function of predicting the target words:
\begin{align}
    \mathcal{L}_{s}=-\textstyle \sum_{t=1}^{L_y} \log P_t^{\text {vocab}}\left(y_{t}\right).
\end{align}

\subsection{Contrastive Learning}

\begin{figure}
    \centering
    \includegraphics[width=1\columnwidth]{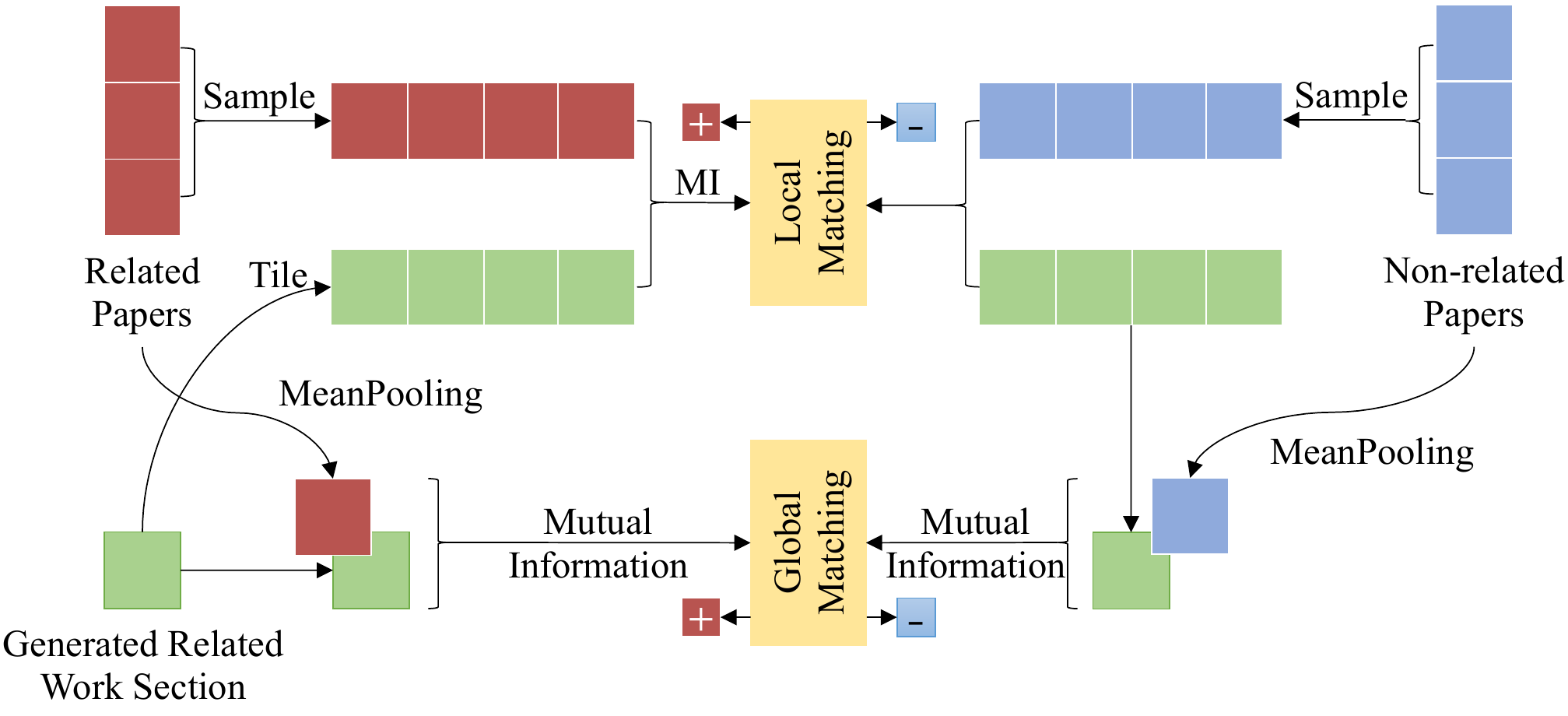}
    \caption{
          The framework of the contrastive learning module.  Mutual Information (MI) is applied to measure both global and local matching. 
    }
    \label{fig:contrastive}
\end{figure}

As we found in preliminary experiments, the ``safe response problem'' \cite{ma2021one} also exists in related work generation task, which means that the models are likely to generate templated sentences without concrete information about the related works.
Thus, to generate more accurate related work sections that are consistent with the detailed facts in the references, we add a contrastive learning module on top of the sequence-to-sequence architecture to provide additional training signals.
Naturally, we expect the relatedness of the generated text with the related paper references to be higher than that with the non-related works. 
In this way, we can avoid generating templated words that cannot distinguish the relevent papers and irrelevant papers.
Following \citet{hjelm2018learning}, we employ the mutual information estimator to estimate such relatedness.
An overview of the module is shown in Figure \ref{fig:contrastive}.

We first use a local matching network for the mutual information measured on the level of individual paper samples and the generated text.
From the graph initialization in  section \ref{sec:paper-encoder}, one reference paper is represented at word-level as $[h^{N_1}_{w_{i,1}},...,h^{N_1}_{w_{i,L_r}}]$, and we omit the iteration index from now on for brevity.
Non-related paper is similarly represented as $[\bar{h}_{w_{i,1}},...,\bar{h}_{w_{i,L_f}}]$. 
The final decoder state of generator  $g_{L_y}$ is concatenated with the word-level paper representation to calculate the local matching scores:
\begin{align}
    \tau^r_l = \text{SCORE}([g_{L_y}; {h}_{w_{i,1}}]...[ g_{L_y} ;{h}_{w_{i,L_r}}]), 
\end{align}
where $\text{SCORE}$ is implemented as a convolutional neural network followed by a feed-forward layer to output $\tau^r_l \in [0,1]$, indicating the matching degree between the generated text with the related works.
$\tau^f_l$ is obtained similarly.
The optimization objective of the local matching network is to minimize:
\begin{equation}
    \mathcal{L}_{local} = - \left( \log(\tau^r_l) + \log(1-\tau^f_l) \right).
\end{equation}

We also have a global matching network to measure the global mutual information between the generated text with the references, as well as non-related works.
By applying mean pooling  on the paper encoding vectors of all references and non-related works, we obtain $h_d$ and $\overline{h}_{d}$.
Then, we employ a feed-forward layer below:
\begin{align*}
    \tau^r_g = \text{FFN}([g_{L_y};h_{d}]),
    \tau^f_g =  \text{FFN}([g_{L_y};\overline{h}_{d}]),
\end{align*}
which outputs $\tau^r_g \in [0,1]$ and $\tau^f_g \in [0,1]$, indicating the global mutual information between the generated text with the related works and with non-related works, respectively.
The objective of this global matching network, similar to the local matching network, is to minimize:
\begin{equation}
    \mathcal{L}_{global} = - \left( \log(\tau^r_g) + \log(1-\tau^f_g) \right).
\end{equation}
Finally, we combine the local and global loss functions to obtain the overall loss $\mathcal{L}$, whose gradients are used for updating all model parameters:
\begin{equation}
    \mathcal{L} =  \mathcal{L}_{global} + \mathcal{L}_{local} + \mathcal{L}_s. \label{total-loss}
\end{equation}

\section{Experiments}

\subsection{Datasets}

To evaluate our proposed method in related work generation, we collect the target paper-aware related work generation datasets from the public scholar corpora S2ORC \cite{lo2020s2orc} and Delve \cite{akujuobi2017delve}, named as TAS2 and TAD, respectively. 
TAS2 consists of scientific papers from multiple domains and TAD focuses on the computer science domain, respectively.
For each case, the generation target is a paragraph in the related work section with more than two citations, as a comprehensive related work usually compares multiple works under the same topic.
The abstract of each reference and the target paper are regarded as input, considering that the main idea of a paper is described in its abstract. 
To ensure the quality of the datasets, we ask three PhD students to evaluate 60 different randomly sampled cases from the datasets.
They are asked to find the original paper content from the scholar websites and to check if the case content is consistent with the original paper.
The result shows that out of 60 sampled cases, 57 are the same as the original paper.
For the left cases, two related work sections of two cases are scattered in the original related work sections, and one abstract section of a case contains part of the introduction section content.
This might be due to the update of the original paper, or mis-segmentation on the paper.

\begin{table}[t]
\centering
\small
\begin{tabular}{l l c c c} \Xhline{2\arrayrulewidth}
\textbf{Dataset} & \specialcell{\textbf{\# Pairs} }&  \specialcell{\textbf{\# words} \\\textbf{(references)}} & \specialcell{\textbf{\# words} \\\textbf{(related work)}} &  \specialcell{\textbf{\# words} \\\textbf{(target)}}
\\ \Xhline{\arrayrulewidth}
RWS \cite{chen2016summarization} & 25 & 5,496 &  367 & 130\\
NudtRwG \cite{wang2019toc} & 50 & 594 &  133 & 147\\
S2ORC \cite{chen2021capturing} & 136,655 & 1,079 & 148 & - \\
Delve \cite{chen2021capturing} & 78927 & 626 & 181 & - \\
TAS2 (ours) & 117,700 & 788 & 126 & 190\\
TAD (ours) & 218,255 & 845 & 191 & 183 \\
\Xhline{2\arrayrulewidth}
\end{tabular}
\caption{Comparison of our datasets to other related work and multi-document datasets. }
\vspace{-10mm}
\label{tab:statistics}
\end{table}

Overall, TAS2 consists of 107,700 training cases, 5,000 validation cases, and 5,000 test cases.
TAD consists of 208,255 training cases, 5,000 validation cases, and 5,000 test cases.
On average, there are 4.26 references in TAS2 and 5.10 in TAD.
Detailed statistics of the datasets and comparison with other related work datasets are given in Table \ref{tab:statistics}.
It can be seen that our datasets are the first large-scale related work dataset with the target paper.

\subsection{Baselines}
We evaluate our models against popular and state-of-the-art multi-document summarization baselines and related work generation techniques.
We adopt the LEAD baseline which selects the first sentence of each reference as the summary as a baseline, and ORACLE as an upper bound of extractive summarization systems.

Our extractive baselines include: 

\textbf{LexRank} \cite{erkan2004lexrank}: a graph-based extractive model that computes sentence importance based on the concept of eigenvector centrality in a graph representation of sentences. 

\textbf{NES} \cite{wang2018neural}: a neural data-driven summarizer that extracts sentences from reference papers based on target paper.
A joint context-driven attention mechanism is proposed to measure the contextual relevance within full texts.

\textbf{BertSumEXT} \cite{Liu2019TextSW}: an extractive summarization model with BERT encoder, which is able to express the semantics of a document and obtain representations for its sentences. 

Abstractive baseline models include: 

\textbf{BertSumABS} \cite{Liu2019TextSW}: an abstractive summarization system built on BERT with a new fine-tuning schedule that adopts different optimizers for the encoder and the decoder.

	\begin{table*}[htb]
	\small
		\centering
		\begin{tabular}{l cccc cccc}
			\toprule
			\multirow{3}{*}{Models} & \multicolumn{4}{c}{\textsf{TAS2 Dataset}} & \multicolumn{4}{c}{\textsf{TAD Dataset}} \\
			\cmidrule(lr){2-5} \cmidrule(lr){5-9} 
			& ROUGE-1 & ROUGE-2 & ROUGE-L & ROUGE-SU & ROUGE-1 & ROUGE-2 & ROUGE-L & ROUGE-SU \\
			\midrule
			\textit{ORACLE} & 37.50 & 7.99 & 32.49 &12.65 & 37.89 & 8.53 & 33.28 \\
			\midrule
			\multicolumn{4}{@{}l}{\emph{Sentence extraction methods}}\\
		    LEAD & 19.63 & 1.75 & 16.88 & 3.73 & 22.74 & 2.32 & 20.15 & 4.87 \\
			LexRank ~\cite{erkan2004lexrank} & 25.74 & 2.81 & 22.43 &6.03 & 25.70 & 2.86 & 22.68 &5.90\\
			LexRank \textit{w.T} \cite{erkan2004lexrank} & 27.04 & 3.18 & 23.48 & 6.55  & 27.29 & 3.50 & 24.06 & 6.61\\
			NES \textit{w.T} \cite{wang2018neural} & 26.04 & 3.39 & 22.46 & 6.14 & 26.13 & 3.24 & 23.20 & 6.18\\
			BertSumEXT ~\cite{Liu2019TextSW} & 25.85 &2.90 & 22.66 & 6.21& 25.95 & 2.92 & 23.05 & 6.25 \\
			BertSumEXT \textit{w.T} \cite{Liu2019TextSW} & 27.43 & 3.56 & 24.01& 6.97 & 27.60 & 3.64 & 24.51 & 6.98\\
			\midrule
			\multicolumn{4}{@{}l}{\emph{Abstractive methods}}\\
            BertSumABS~\cite{Liu2019TextSW} & 25.45 & 3.82 & 23.04 & 6.39 & 27.42 & 4.88 & 25.15 & 7.22\\
            MGSum~\cite{Jin2020MultiGranularityIN} & 25.54 & 3.75 & 23.16 & 6.49 & 27.49 & 4.79 & 25.21 & 7.29\\
            EMS~\cite{zhou2021entity} & 26.17 & 4.16 & 23.63 & 6.67 & 28.21 & 5.15 & 25.74 & 7.56 \\
            EMS \textit{w.T} & 26.50 & 4.22 & 23.90 & 6.84 & 28.78 & 5.36 & 26.37 & 7.89\\
            RRG~\cite{chen2021capturing} & 26.31 & 4.41 & 24.05 & 6.63 & 28.44 & 5.43 & 25.97 & 7.65\\
            RRG \textit{w.T} & 26.79 & 4.43 & 24.46 & 6.85 & 28.94 & 5.59 & 26.46 & 7.92 \\
            TAG & \textbf{28.04} &\textbf{4.75} & \textbf{25.33}  & \textbf{7.69} & \textbf{30.48} & \textbf{6.16 }& \textbf{27.79} & \textbf{8.89}\\
             \midrule
             \multicolumn{4}{@{}l}{\emph{Ablation models}}\\
             TAG w/o graph encoder & 27.03 & 4.53 & 24.47 & 7.21 & 29.34 & 5.76 & 26.85 & 8.32 \\
             TAG w/o hierarchical decoder & 27.45 & 4.66 & 24.84 & 7.38& 29.64& 5.88 & 27.07 & 8.43\\ 
             TAG w/o contrastive learning & 27.27 & 4.55 & 24.72 & 7.33 & 29.48 & 5.46 & 26.93 & 8.37 \\ 
			\bottomrule
		\end{tabular}
		\caption{ ROUGE scores comparison between TAG and baselines.
		All our ROUGE scores have a 95\% confidence interval of at most $\pm$0.24 as reported by the official ROUGE script.
		\textit{w.T} denotes \textit{with.Target} as input.}
		\label{tab:baselines}
	\end{table*}

\textbf{MGSum} \cite{Jin2020MultiGranularityIN}: a multi-granularity interaction network for abstractive multi-document summarization.
The multi granularities include document-level, sentence-level, and word-level.

\textbf{EMS} \cite{zhou2021entity}: an entity-aware model for abstractive multi-document summarization with BERT encoder.
It augments the classical Transformer based encoder-decoder framework with a graph consisting of paragraph nodes and entity nodes.
EMS uses the same keyphrase extraction tool as ours.

\textbf{RRG} \cite{chen2021capturing}: an abstractive related work generator based on reference papers.
It augments the RNN encoder with a relation-aware graph between multiple references.
The relation graph and the document representation are refined iteratively.

The essential difference between our model and baselines is that we explicitly model the relationship between the target work and the references.
To make a fair comparison, for baselines that have relatively good performance, we give an additional \textit{with.Target} (\textit{w.T}) version which takes references and target paper as input.
Thus, the baseline models treat the references and the target paper as multiple inputs without distinguishing them.

\subsection{Evaluation Metrics}
For both datasets, we evaluate standard ROUGE-1, ROUGE-2, ROUGE-L, and ROUGE-SU~\cite{lin2004rouge} following \cite{chen2018iterative} on full-length F1, which refer to the matches of unigram, bigrams, the longest common subsequence, and skip bigrams with a max distance of four words, respectively.

\citet{schluter2017limits} notes that only using the ROUGE metric to evaluate generation quality can be misleading. 
Therefore, we also evaluate our model by human evaluation.
Since our datasets are scientific corpus with complicated concepts and rare, technical terms, the human evaluation should be conducted by professional annotators.
Hence, we ask three PhD students majoring in computer science to rate 40 randomly sampled cases generated by models from the TAD dataset, since TAD consists of computer science papers.
The evaluated models are EMS \textit{w.T}, RRG \textit{w.T}, and TAG, which achieve top performance in automatic evaluations.

Our first evaluation quantified the degree to which the models can retain the key information following a question-answering paradigm \cite{Liu2019HierarchicalTF}.
	We created a set of questions based on the gold-related work and examined whether participants were able to answer these questions by reading generated text. 
	The principle for writing a question is that the information to be answered is about factual description, and is necessary for the related work. 
	Two PhD students majoring in computer science wrote three questions independently for each sampled case. 
	Then they together selected the common questions as the final questions that they both consider to be important. 
	Finally, we obtain 68 questions, where correct answers are marked with 1 and 0 otherwise. 
	Our second evaluation study assessed the overall quality of the related works by asking participants to score them by taking into account the following criteria: \textit{Informativeness} (does the related work convey important facts about the topic in question?), \textit{Coherence} (is the related work coherent and grammatical?), and \textit{Succinctness} (does the related work avoid repetition?). 
	The rating score ranges from 1 to 3, with 3 being the best. 
	Both evaluations were conducted by another three PhD students independently, and a model’s score is the average of all scores.

\subsection{Implementation Details}
The Transformer in the graph initialization module is initialized by BERT base model \cite{devlin2019bert}, which can be more effective than that trained from scratch.
We also equip baselines with BERT  (if applicable) for better comparison.
Both source and target texts were tokenized with BERT's subwords tokenizer.
We use hyperparameters suggested by \citet{Liu2019TextSW}.
The iteration numbers $N_1$, $N_2$, and $N_3$ are set to 6, 2, 6.
The hidden size is 768 and the number of heads is 6, while the hidden size of feed-forward layers is 1,024. 
We use mini-batches of size 16, dropout with probability 0.1 before all linear layers, and label smoothing \cite{szegedy2016rethinking} with smoothing factor 0.1.
For one generation, 5 reference papers and 20 keyphrases are considered (with padding if not enough).  
For abstractive models, the first 200 words in the abstract are taken as paper content.
For extractive baselines, the first 10 sentences with a maximum of 20 words per sentence in the abstract are taken as paper content.
During decoding, we set the minimum decode length to 100 tokens, and the maximum length to 150 tokens.
We use Adam optimizers with learning rate of 0.0001.
We apply beam search with beam size 5 and length penalty \cite{wu2016google} with factor 0.4.

\subsection{Experimental Results
}

\begin{figure*}[htb]
\begin{boxedminipage}{2\columnwidth}
\scriptsize
\textbf{Reference Papers (trancated):}
\newline
[2] \fph{Distributed hash table} (DHT) based overlay networks offer an administration-free and fault-tolerant storage space that maps ``keys'' to ``values''.
For these systems to function efficiently, their structures must fit that of the underlying network.
Existing techniques for discovering network \reph{proximity information}, such as landmark clustering and expanding-ring search are either inaccurate or expensive.
\newline
[3] \fph{Distributed hash table} (DHT)-based overlay networks, represented by Pastry, CAN, and Chord, offer an administration-free and fault-tolerant application-level overlay network.
While elegant from a theoretical perspective, these systems have some disadvantages.
First, they rely on application level routing, which may be inefficient with respect to network delays and bandwidth consumption.
Second, they typically construct a homogeneously structured overlay even though nodes in these networks usually have varying physical connectivity and packet-forwarding capacities.
\newline
\textbf{Target Paper (truncated):}

Many solutions have been proposed to tackle the problem of load balancing in structured P2P systems.
However, all these solutions either ignore the heterogeneity nature of the system, or reassign loads among nodes without considering proximity relationships, or both.
In this paper, we propose an efficient scheme to ensure fair load distribution in such P2P systems by utilizing \reph{proximity information}.
To our knowledge, this is the first work to utilize the proximity information to guide \error{load balancing}.\\
\textbf{QA}: 
What information has been used for DHT construction?  [Proxitmity information]\\
What is the task of this work? [Load balance]
\newline \rule{\columnwidth}{0.4pt}
\textbf{Ground Truth Related Work:}
\reph{Proximity information} has been utilized in both topologically aware \fph{DHT}  construction [1] and proximity neighbor selection in P2P routing tables [2] [3].
The primary purpose of using the proximity information in both cases is to improve the performance of DHT overlays. 
However, the proximity information used in our work is to make \error{load balancing} fast and efficient.
\newline \rule{\columnwidth}{0.4pt}
\textbf{Baseline EMS \textit{with.Target}:}
In [1], the authors propose a \fph{distributed hash table (DHT)} scheme that uses \reph{proximity information} to improve the performance of a DHT system.
In this scheme, each node maintains a hash table for each node in the network.
The hash table is used to determine the location of the nodes in the system, and the node is responsible for the hashing of the keys.
A DHT-based system is proposed in [2].
This scheme is based on the assumption that the nodes are physically close to each other.
\newline \rule{\columnwidth}{0.4pt}
\textbf{Baseline RRG \textit{with.Target}:}
In this paper, we focus on the analysis of wireless ad-hoc overlay network search, which is based on the \reph{proximity information}. 
In [1], the authors propose a distributed topological model to model overlay networks.
In [2], the authors discuss the problem of overlay networks with \fph{distributed hash table}.
In [3], the authors propose a similar approach.
In this work, we are interested in the context of overlay networks.
\newline \rule{\columnwidth}{0.4pt}
\textbf{Our TAG model:}
Our work is also related to  \reph{proximity information} in P2P systems [1].
In [2], the authors propose to use proximity information to improve the performance of P2P systems.
In [3], authors propose to use proximity information to improve the performance of \fph{distributed hash table (DHT)}-based P2P networks.
However, their work does not address the issue of \error{load balancing} in P2P networks.
\end{boxedminipage}
\caption{
Selected examples of the related work section generated by the baselines and our model.
The same color denotes the same key information.
}
\label{case}
\end{figure*}

\begin{table}[htb]
\centering
\small
\begin{tabular}{@{}lcccc@{}}
\toprule
& QA(\%) & Info & Coh & Succ\\
\midrule
EMS \textit{w.T} & 32.3 & 2.14 & 2.20 & 2.05 \\
RRG \textit{w.T} &36.7 & 2.32 & 2.29 & 2.16 \\
TAG & \textbf{41.1} & \textbf{2.49} & \textbf{2.36} & \textbf{2.28} \\
\bottomrule
\end{tabular}
\caption{Comparison of human evaluation in terms of QA task, informativeness (Info), coherence (Coh) and succinctness (Succ).}
\vspace{-5mm}
\label{tab:human_evaluation}
\end{table}

\textbf{Automatic evaluation.}
A comparison of the performance between TAG and state-of-the-art baselines is listed in Table~\ref{tab:baselines}.
Firstly, it can be seen that for extractive models, adding target paper information can improve their performances by a large margin.
For example, the ROUGE-L score increases by 2.70 on TAS2 dataset for LexRank.
This demonstrates the effectiveness of the target paper information in related work generation task.
However, such improvement is not obvious for abstractive baseline models.
The ROUGE-L score of EMS only improves by 0.27 and the ROUGE-2 score improves by 0.06 on TAS2 dataset.
One potential reason is that it is hard to recognize the contribution of the target without explicitly modeling the relationship between the target paper with the references. 
Finally, our TAG outperforms various previous abstractive models on all metrics, which shows that our model can take advantage of the target paper better.
Specifically, our model improves ROUGE-1 by 4.66\% over RRG \textit{w.T} on the TAS2 dataset, and improves ROUGE-L by 7.00\% on the TAD dataset.
We also conducted significant testing (t-test) on the improvements of our approaches over the best baseline. 
The results indicate that all the improvements are significant (p-value $< 0.05$). 
The above observation demonstrates the necessity of a tailored related work generation model compared with traditional multi-document summrization model.
It also shows the effectiveness of our model on different types of corpora.

\textbf{Human evaluation.}
Table~\ref{tab:human_evaluation} summarizes the results of human evaluation on three baselines and the proposed model. 
TAG is the top performer over all the metrics. 
Specifically, in the QA task, our model answers 4.4\% more questions than RRG \textit{w.T}.
our model outperforms RRG \textit{w.T} by 0.17 on \emph{Informativeness}, suggesting that TAG captures salient information in the references and the target paper.
As for the \emph{Coherence} metric, our model scores 0.07 higher than RRG \textit{w.T}. 
This indicates that TAG manages to reduce grammatical errors and improve the readability of the generated related work.
On the \emph{Succinctness} metric, our model outperforms all baselines by a considerable margin, suggesting that the generated related work avoids redundancy.
Note that the average Kappa statistics of human evaluation are 0.42, 0.45, and 0.41, respectively, indicating moderate agreement between annotators.
The paired student t-test between our model and RRG \textit{w.T} provides further proves of the significance of these results.

\section{ANALYSIS AND DISCUSSION}
\label{sec:analysis}

\begin{figure*}[htb]
\begin{boxedminipage}{2\columnwidth}
\scriptsize
\textbf{Ground Truth Related Work:}
Numerous approaches for content-based image retrieval have been proposed in the literature. 
They are based on features like color [1], shape [2] or texture [3]. 
In [4] a graph-based approach similar to the one described in section 3.2 is used, while in [5] an edit distance measure is used to measure similarity of topological arrangements. 
[6] gives a nice overview of the different approaches. 
However, all those approaches concentrate on single image features like color or texture.
\reph{Our method differs from those proposed in the literature as we combine different approaches in order to achieve more meaningful results.}
\newline \rule{\columnwidth}{0.4pt}
\textbf{Baseline EMS \textit{with.Target}:}
A number of methods have been proposed for image retrieval.
Some of these methods are based on the analysis of the image content [1] [2].
These methods can be classified into two categories: content-based and concept-based methods .
Content based methods, such as [3], use a combination of content and content features.
The content of the images is then used to determine the similarity between images.
\error{However, these methods do not deal with the problem of image segmentation}.
\newline \rule{\columnwidth}{0.4pt}
\textbf{Baseline RRG \textit{with.Target}:}
There has been a number of work in the field of image retrieval [1]. 
In [2], the authors propose an image retrieval system based on the content of the images.
The system is based on a combination of color and texture features.
A similar approach is used in [3] for image retrieval based on color, texture, and shape features.
\error{However, they are not directly applicable to image databases because they are limited to a single image, which is not the case in our case.}
\newline \rule{\columnwidth}{0.4pt}
\textbf{Our TAG model:}
Content-based image retrieval (cbir) has been studied extensively in the last decade.
Many approaches have been proposed for cbir. 
Some of them are based on color, texture , and shape features [1][2][3].
Others use color and texture features to improve the performance of cbir systems.
For example, the qbic system [4] uses color, shape, and texture as features for image retrieval. 
In [5], the authors use shape features for cbir.
\error{But they do not consider the spatial relationship between objects}.
\end{boxedminipage}
\caption{
Error analysis case of our model and baselines.
Red denotes error information.
}
\label{case3}
\end{figure*}

\subsection{Ablation Study}

\begin{figure}
    \centering
    \includegraphics[width=0.9\columnwidth]{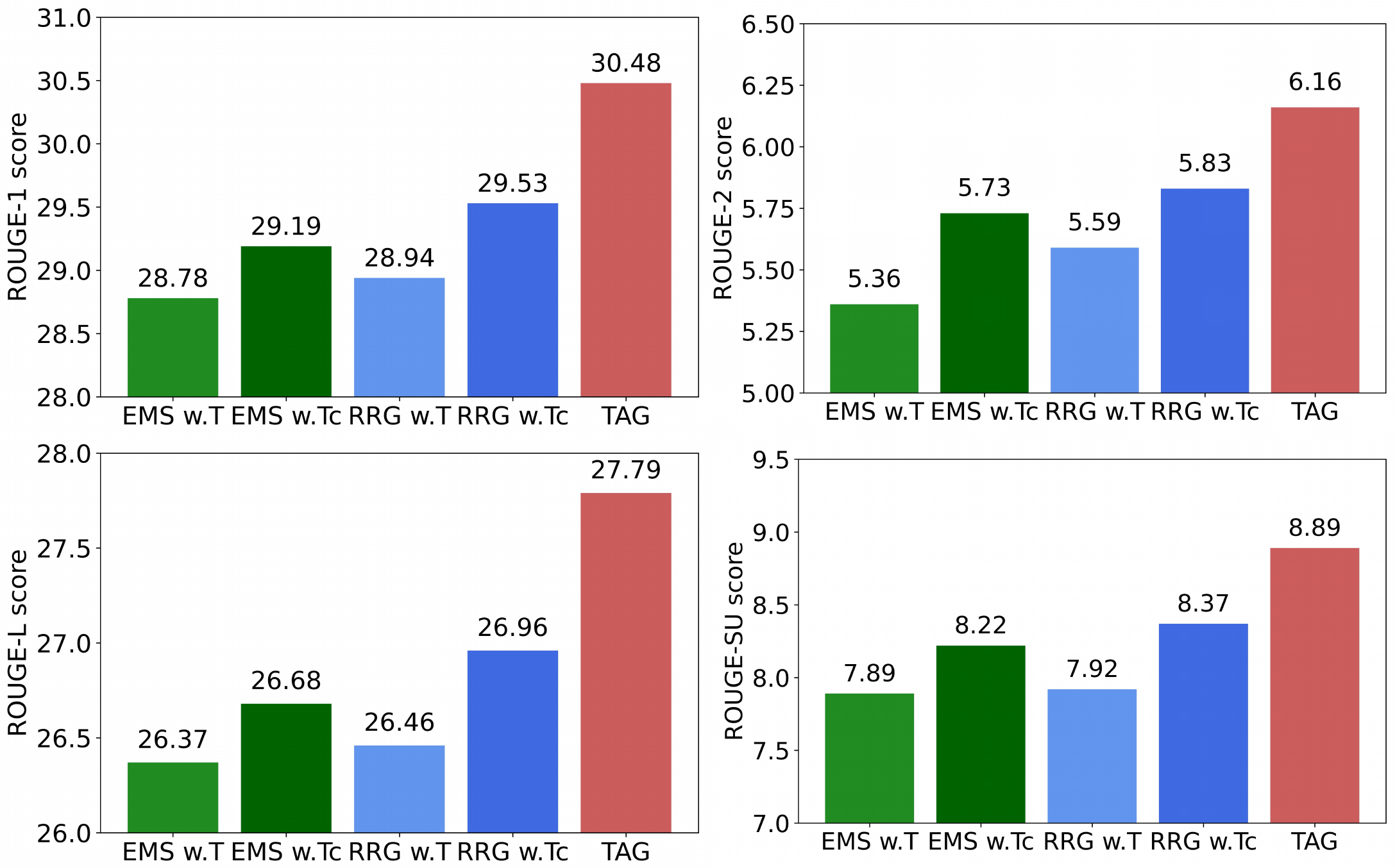}
    \caption{
Performances of baseline models equipped with our contrastive learning module. 
    }
    \label{contras}
\end{figure}

We perform an ablation study on the development set to investigate the influence of different modules in our proposed TAG model. 
Modules are tested in four ways:
(1) we remove the target-aware graph encoder to verify the effectiveness of the graph encoder (w/o graph encoder).
Since there are no interactions between different documents without the graph encoder, we first obtain the document-level representations by taking the mean pooling of the word-level representations, and then apply cross-attention mechanism on these document representations.
(2) we replace the hierarchical decoder with a naive Transformer decoder without the keyphrase-centered mechanism, to verify the effectiveness of the guidance of keyphrases (w/o graph encoder).
(3) we remove the contrastive learning module to verify the effectiveness of the importance of non-related works (w/o contrastive learning). 

The last block of Table~\ref{tab:baselines} presents the results.
We find that the ROUGE-1 score drops by 1.14 after the graph encoder is removed on TAS2 dataset.
The results confirm that employing the target-centered graphic interaction can generate better related work.
ROUGE-L score drops by 0.72 after the hierarchical decoder is removed on TAG dataset.
This indicates that the keyphrases do help the model identify high-level gist and guide the generation process.
Last, there is a substantial drop in all metrics when the contrastive learning module is excluded.
This demonstrates that introducing the information from negative samples can help the model generate text more related to the positive samples.

\subsection{Effectiveness of add-on Contrastive Learning}

Note that our contrastive learning module is an add-on module, since it only takes the final representations of the inputs and the generated text, and does not have requirements on the internal structure of the model.
Hence, we equip baselines with the contrastive learning module, to verify if our contrastive learning is effective in different scenarios.
We choose EMS \textit{w.T} and RRG \textit{w.T} as baselines since they perform better than other baselines.

The comparison result on TAS2 dataset is shown in Figure~\ref{contras}.
It can be seen that with contrastive learning, both of EMS \textit{w.T} and RRG \textit{w.T}  perform better substantially.
Concretely, the ROUGE-1 score of EMS \textit{w.T} increase from 28.78 to 29.19, and the ROUGE-L score of RRG \textit{w.T} increases from 26.46 to 26.96.
This demonstrates the universality of our model with different generation structures.
Besides, even equipped with the contrastive module, both baselines still underperform our TAG.
This proves the superiority of our main generation model, which consists of a target-aware graph encoder and a target-aware decoder.

\subsection{Case Study}

Figure~\ref{case} presents an example of the related work section generated by TAG and baseline models. 
We observe that most of the generated related work can capture the high-level key information in the references such as ``proximity information'' can improve the performance of ``distributed hash table'' construction (highlighted in color).
However, most of the models except our TAG cannot address the ``load balancing problem''.
One possible reason is that this keyphrase obtains a refined representation output by the graph encoder, and then is recognized by the hierarchical decoder.

Next, we give an error analysis in Figure~\ref{case3}.
It can be seen that this case is about the image retrieval task, and most of the generated related works can give this key information and introduce that color, shape, and texture features can be used to improve the performance of the task.
However, when introducing the difference between target work and related works, all models make some mistakes.
EMS \textit{with.Target} said that the contribution of the target work is to deal with the \underline{image segmentation problem}.
RRG \textit{with.Target} said that existing works can only be applied to a \underline{single image}.
Our model describes that the target work is the first work \uline{considers spatial relationship between objects}.
In fact, the ground truth contribution of the target paper is that it  \uline{combines different approaches including spatial features and textural features}.
Unfaithful problem is a common phenomenon in summarization models, and we are looking forward to bringing more advanced generation models.

\subsection{Visualization Analysis}

\begin{figure}[htb]%
    \centering
    \subfigure{{\includegraphics[width=3.7cm]{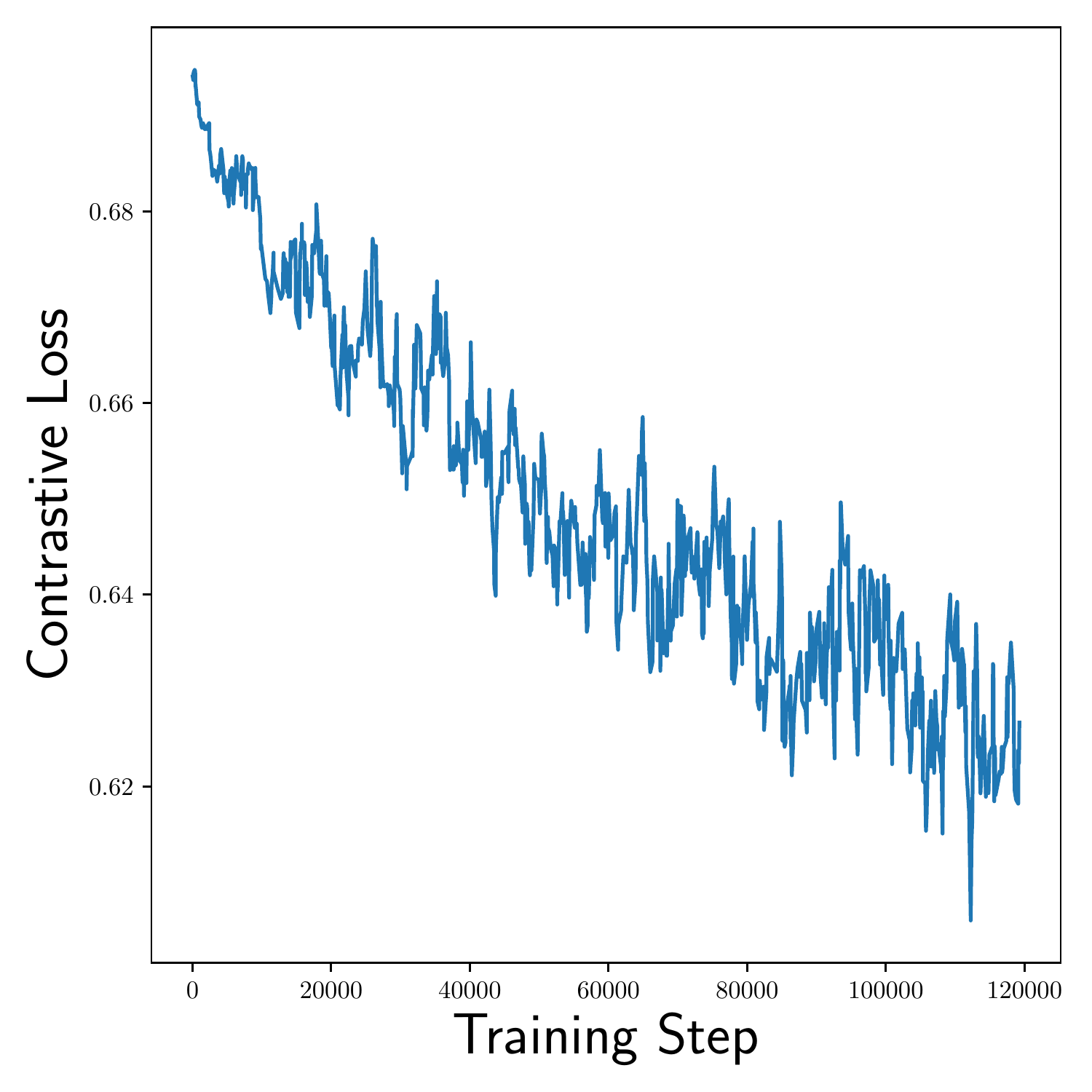} }}%
    \subfigure{{\includegraphics[width=4.8cm]{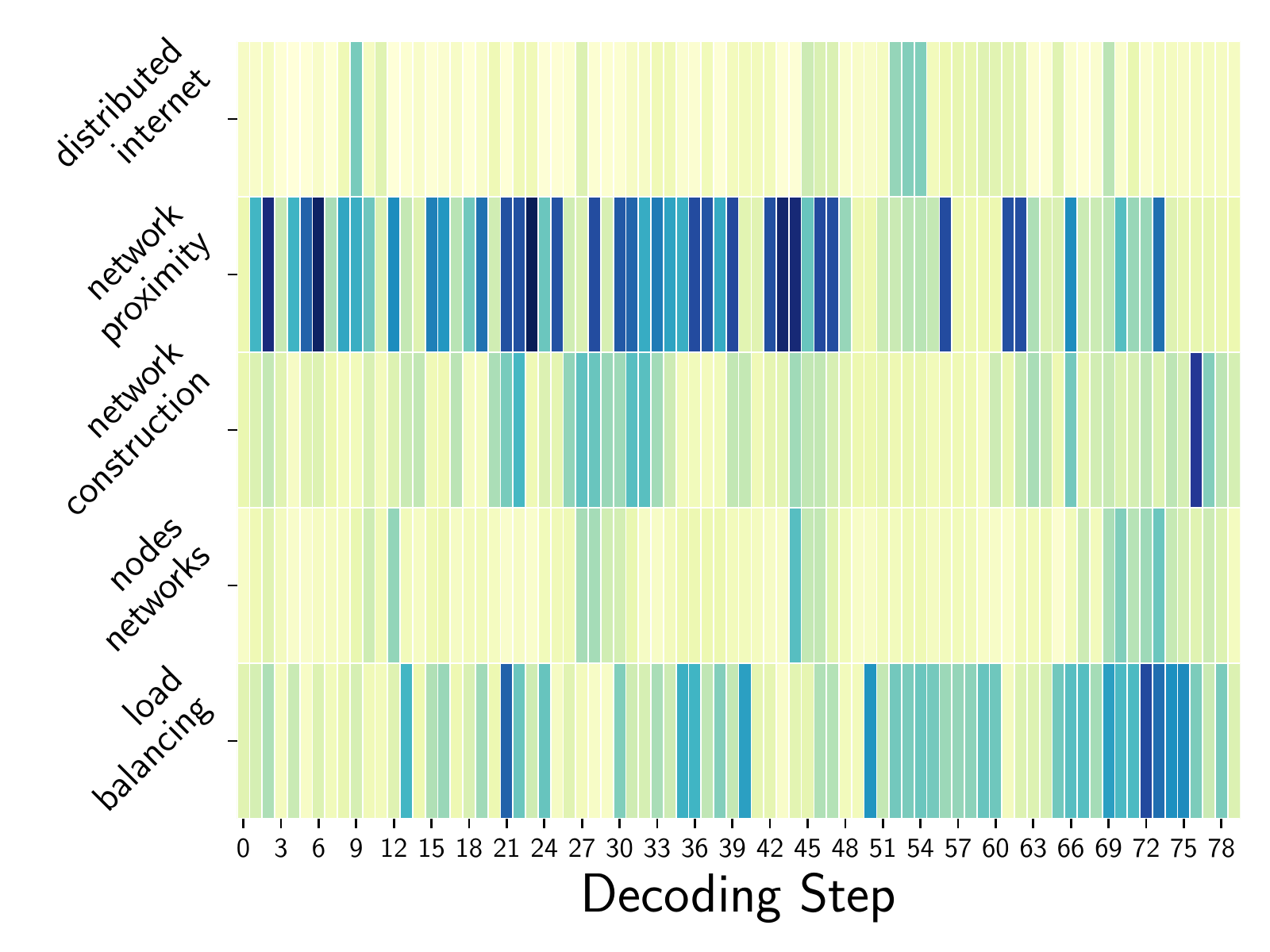} }}%
    \caption{(a) Loss curve of the matching network in the contrastive learning module. (b) Visualizations of the attention weights on references.}%
    \label{f2}%
\end{figure}

In this subsection, we give visualization analysis on the contrastive learning process and the decoding process.
We first visualize the loss curve of the local matching network in Figure~\ref{f2}(a).
It shows that the loss score fluctuates at the beginning of the training and gradually reaches convergence along with the training.
This phenomenon demonstrates that the matching network can gradually identify the generated text with the related work, and mutual information is gradually maximized. 
The accuracy curve of the global matching network is similar and is omitted here.
Besides, we visualize the attention weights $z_{c,t}^{N_3}$ on keyphrases for the case in Figure~\ref{case}.
The x-axis is the decoding step and the y-axis is the selected keyphrases.
It can be seen that when generating the first several sentences, the keyphrase ``network proximity'' wins higher attention than others.
It thus successfully leads the generated related work to describe the related works on ``proximity information''.
At the end of the generation, the keyphrase ``load balancing'' obtains higher attention.
Subsequently, the generated text describes the key difference between the target work and the related works.

\section{Conclusion}

In this paper, we propose to incorporate the target paper information into the abstractive related work generation task.
Concretely, we propose a target-centered graph encoder and a target-guided hierarchical decoder on top of the Transformer framework.
To generate a more substantial related work, we adapt the contrastive learning mechanism to our model.
Experimental results demonstrate that our model outperforms the latest baselines by a large margin in automatic and tailored human evaluations.
In the future, we target to bring a multi-lingual related work generation system.

\section*{Acknowledgments}
We would like to thank the anonymous reviewers for their constructive comments. 
The work was supported by National Key R\&D Program of China (2020YFB1406702), National Natural Science Foundation of China (NSFC Grant No. 62122089 and No. 61876196), and King Abdullah University of Science and Technology (KAUST) through grant awards Nos. BAS/1/1624-01, FCC/1/1976-18-01, FCC/1/1976-23-01, FCC/1/1976-25-01, FCC/1/1976-26-01.
\bibliographystyle{ACM-Reference-Format}
\bibliography{sample-base}

\end{document}